\documentclass[twocolumn]{fairmeta}
\usepackage[most]{tcolorbox}

\usepackage{amsmath}
\usepackage{amssymb}
\usepackage{capt-of}

\title{Unified Condition-Action Modeling for Accurate One-Step Action Generation}

\author[1,*]{Xinyu Zhou}
\author[1,*]{Zikun Cai}
\author[1]{Kuangji Zuo}
\author[1]{Gen Li}
\author[1]{Boyu Ma}
\author[1]{Yanshuo Lu}
\author[1]{Yutong Song}
\author[2]{Mingqi Yuan}
\author[2,\dagger]{Jiayu Chen}
\author[1,\dagger]{Jianfei Yang}

\affiliation[1]{MARS Lab, Nanyang Technological University}
\affiliation[2]{The University of Hong Kong}

\contribution[*]{Equal Contribution}
\contribution[\dagger]{Corresponding Authors}

\abstract{
    Robot manipulation requires policies that are both accurate and efficient, as robot control must respond to changing observations under tight latency constraints. Recent diffusion and flow policies are promising, but they often treat conditions as auxiliary signals rather than jointly evolving them with action trajectories. We find that this limitation can be effectively mitigated by a \textbf{simple yet effective unified condition-action modeling design} that represents conditions and actions in a shared token space, allowing a compact model to achieve high performance while improving both inference speed and accuracy. Therefore, we propose UCA-Flow, a unified condition-action modeling framework for accurate one-step action generation. Our method unifies observation conditions, timestep conditions, interval conditions, and action tokens into a single sequence, and processes them with a Unified Condition-Action Transformer for joint condition-action representation learning. As a result, condition representations are dynamically reconstructed according to the current generation stage, highlighting information most relevant for action refinement. Furthermore, we introduce an improved dual-pass supervision scheme over $u$ and $v$ for stronger optimization of unified condition-action modeling.
    UCA-Flow achieves an average success rate of $86.4\%$, improving over the strongest baseline by 7.5 percentage points, with a task-averaged inference latency of $3.0$ ms. It achieves $43.9\times$ and $32.2\times$ speedups over DP3 and Simple DP3, and remains $4.2\times$ and $2.2\times$ faster than one-step FlowPolicy and MP1, respectively. Project page: \url{https://uca-policy.github.io/UCA.github.io/}.
}

\correspondence{Jianfei Yang at \email{jianfei.yang@ntu.edu.sg}, Jiayu Chen at \email{jiayuc@hku.hk}}

\begin{document}

\maketitle

\section{Introduction}

Robot manipulation aims to generate executable actions from complex observations for physical interaction, such as grasping, pushing, assembly, tool use, and dexterous object manipulation \cite{suomalainen2022survey,bohg2014data,qin2023robot}, with broad applications in industrial automation, service robotics, and logistics\cite{suomalainen2022survey,xu2024survey}. A key challenge is to extract task-relevant information while generating actions under tight control latency for closed-loop execution. Recent generative robot policies, including Diffusion Policy\cite{chi2025diffusion}, Flow Matching\cite{zhang2025flowpolicy}, and Mean Flow\cite{sheng2025mp1meanflowtamespolicy}, have shown strong potential for modeling action distributions \cite{chi2025diffusion, ze20243ddiffusionpolicygeneralizable, zhang2025flowpolicy, sheng2025mp1meanflowtamespolicy, yan2025maniflow}. However, they still face an efficiency-performance bottleneck.

\begin{figure*}[tbhp]
  \centering
  \includegraphics[width=\textwidth]{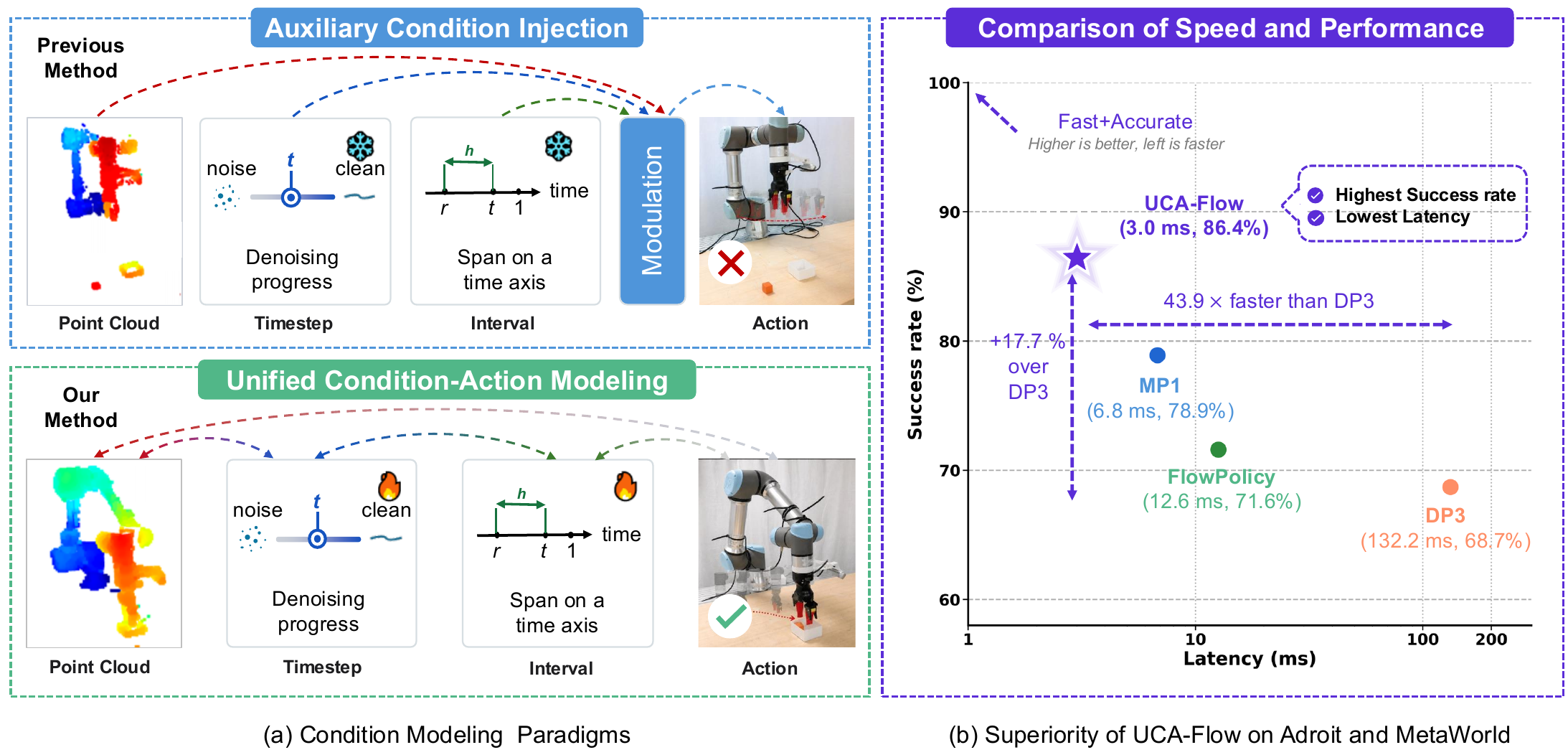}
\caption{
\textbf{Motivation of UCA-Flow}. (a) Existing policies often treat conditions as auxiliary signals. Unified condition-action modeling enables conditions and actions to evolve together, allowing condition representations to adapt to the current generation stage and action. (b) UCA-Flow achieves $86.4\%$ average success and $3.0$ ms task-averaged inference latency on 37 Adroit and Meta-World tasks, improving success over DP3 by 17.7 percentage points with a $43.9\times$ speedup. Speedups use task-weighted latencies before rounding.
}
\label{fig:motivation}
\end{figure*}

To address this challenge, prior work has improved generative robot policies from several complementary directions. Some methods introduce compact 3D point-cloud features or multimodal state embeddings as conditions for action generation, improving spatial understanding and data efficiency \cite{ze20243ddiffusionpolicygeneralizable, zhang2025flowpolicy, sheng2025mp1meanflowtamespolicy}. Others improve sampling efficiency through flow matching, consistency training, or mean-flow objectives, enabling few-step or one-step action generation \cite{zhang2025flowpolicy, geng2025mean, sheng2025mp1meanflowtamespolicy}. Recent transformer-based policies further enhance multimodal conditioning through cross-attention, adaptive cross-attention, AdaLN, or AdaLN-Zero modulation, allowing action tokens to selectively attend to visual, language, timestep, and proprioceptive inputs \cite{reuss2024multimodal, yan2025maniflow}. Despite these advances, as illustrated in the upper part of Figure \ref{fig:motivation}(a), most designs still follow a condition-to-action paradigm: observations and timesteps are encoded as external auxiliary signals and injected into the action-generation backbone through modulation, rather than represented as first-class tokens that dynamically interact with the evolving action. This condition-injection bottleneck is especially restrictive for one-step action generation.

In this paper, we propose UCA-Flow, a unified condition-action modeling framework for accurate one-step action generation. As shown in the lower part of Figure \ref{fig:motivation}(a), instead of injecting conditions into an action generator as external guidance, UCA-Flow represents observation conditions, timestep conditions, interval conditions, and action tokens as a unified token sequence. The sequence is processed by a compact Unified Condition-Action Transformer, allowing condition and action representations to be jointly refined within a unified representation space. Thus, condition reconstruction becomes an internal part of generation rather than a pre-computed external input. To better optimize unified condition-action modeling, we further introduce an improved dual-pass supervision scheme over average velocity $u$ and instantaneous velocity $v$. Rather than serving as an isolated MeanFlow objective, this dual supervision regularizes the shared condition-action representation from two complementary velocity views, encouraging more stable and discriminative token representations \cite{geng2025mean, geng2025improvedmeanflowschallenges}.
As summarized in Figure \ref{fig:motivation}(b), experiments show that UCA-Flow achieves a substantially improved efficiency-performance trade-off. It improves the average success rate by 7.5 percentage points over the strongest baseline, while achieving $43.9\times$ and $32.2\times$ speedups over DP3 and Simple DP3, and remaining $4.2\times$ and $2.2\times$ faster than one-step FlowPolicy and MP1, respectively.

Our contributions are summarized as follows: 
\begin{itemize}
    \item{We propose UCA-Flow, a unified condition-action modeling framework for accurate one-step action generation, which represents observation, timestep, interval, and action as first-class tokens in a Unified Condition-Action Transformer, enabling condition representations to be dynamically refined according to the generation stage.}
    \item{We introduce an improved dual-pass supervision scheme over average velocity $u$ and instantaneous velocity $v$. This training strategy provides complementary velocity-level constraints to better optimize the unified condition-action modeling framework, improving one-step action accuracy without increasing inference cost.}
    \item {Extensive experiments on 37 Adroit and Meta-World tasks and 2 real-world manipulation tasks demonstrate that UCA-Flow achieves a superior efficiency and performance.}
\end{itemize}

\section{Related works}

\paragraph{Generative policies for robot manipulation.}
Robot manipulation requires policies that infer task-relevant information from complex observations and generate accurate actions for physical interaction~\cite{suomalainen2022survey,bohg2014data,qin2023robot,xu2024survey}. 
Imitation learning provides a practical way to acquire such skills from demonstrations~\cite{atkeson1997robot,argall2009survey,mandlekar2021matters,florence2022implicit,zhao2023learning,haldar2023teach}. 
Early behavior cloning and implicit-policy methods learn observation-action mappings, but they struggle with multimodal action distributions and high-dimensional control~\cite{florence2022implicit,shafiullah2022behavior,zeng2021transporter,kalashnikov2018qtopt,pari2021surprising,hansen2023pretraining}. 
Recent policies address these limitations by modeling action distributions with expressive models. 
Diffusion Policy formulates visuomotor control as conditional action denoising and shows strong performance across simulated and real tasks~\cite{chi2025diffusion}. 
Subsequent works extend this idea to human-behavior imitation, goal-conditioned diffusion, policy optimization, trajectory planning, grasp-motion optimization and flow-based policy learning~\cite{pearce2023imitating,reuss2023goal,ren2024diffusion,janner2022planning,urain2023diffusionfields,prasad2024consistency}. 
However, diffusion-based policies require iterative sampling and increases inference latency.

\paragraph{3D representations for policy learning.}
Because manipulation is inherently spatial, 3D representations are widely used to improve geometric reasoning and data efficiency. 
Benchmarks such as RLBench provide diverse vision-guided manipulation tasks for evaluating such capabilities~\cite{james2020rlbench}. 
Recent policies have explored different forms of 3D representations, including voxels, multi-view images, feature fields, and point clouds, to better capture spatial structure for manipulation.~\cite{shridhar2023peract,gervet2023act3d,goyal2023rvt,goyal2024rvt2,ze2023gnfactor,yan2024dnact,ke2025threediffuseractor}. 
Point-cloud encoders such as PointNet, PointNet++, and PointNeXt provide compact spatial representations that avoid the high cost of dense voxels~\cite{qi2017pointnet,qi2017pointnetplusplus,qian2022pointnext}. 
DP3 shows that sparse point clouds combined with diffusion policies improve data efficiency, and deployment safety~\cite{ze20243ddiffusionpolicygeneralizable}. 
FlowPolicy and MP1 further use compact 3D point-cloud features for efficient flow-based and action generation~\cite{zhang2025flowpolicy,sheng2025mp1meanflowtamespolicy}. 
Nevertheless, these methods still encode observations as external conditions and inject them into action generators through modulation.

\paragraph{Efficient one-step action generation.}
To reduce sampling cost, recent research has explored more efficient sampling and fast generation, including DDPM, DDIM, score-based modeling, flow matching, consistency models, and MeanFlow~\cite{ho2020ddpm,song2021ddim,song2021score,lipman2023flow,liu2022rectifiedflow,song2023consistency,geng2025mean,geng2025improvedmeanflowschallenges}. 
In robot learning, AdaFlow, FlowPolicy, ManiFlow, and MP1 reduce the number of function evaluations for faster policy inference~\cite{hu2024adaflow,zhang2025flowpolicy,yan2025maniflow,sheng2025mp1meanflowtamespolicy}. 
However, most of them improve the generative objective or sampling process,  conditions injected into the action generator as external context. In contrast, UCA-Flow places observation, timestep, interval, and action into a unified condition-action sequence processed by a Unified Condition-Action Transformer, enabling joint condition-action representation learning and dynamic condition reconstruction for accurate one-step action generation.

\section{Method}
We first present the overall formulation of UCA-Flow and its one-step action generation process. 
We then describe how observation conditions, timestep and interval variables, and noisy action are represented as a unified condition-action token sequence. 
Based on this sequence, we introduce Unified Condition-Action Interaction for jointly updating condition and action representations. Finally, we detail the dual-pass supervision objective for optimizing interval-aware average velocity prediction.

\begin{figure*}
  \centering
  \includegraphics[width=1\textwidth]{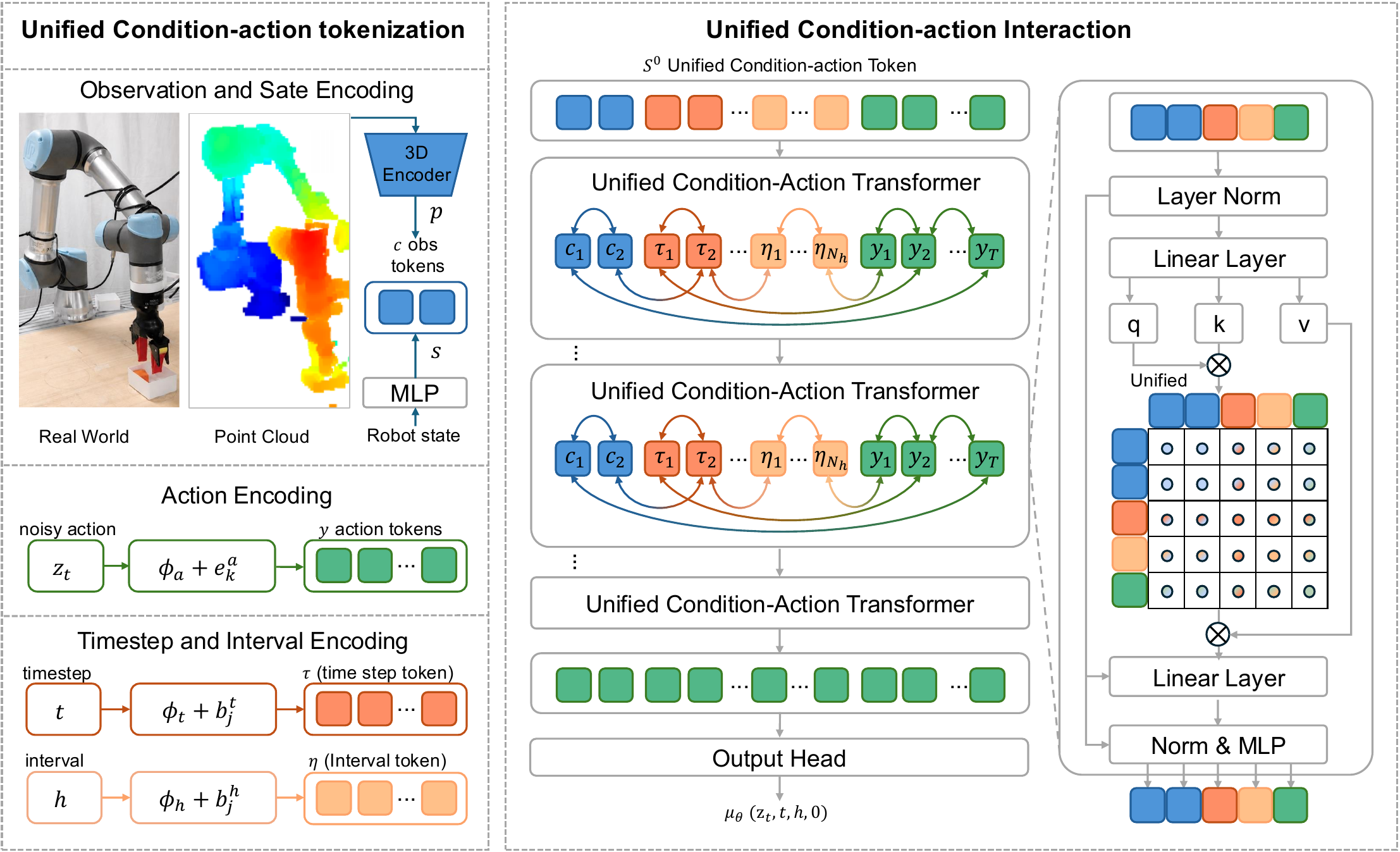}
\caption{
\textbf{Overview of UCA-Flow}. Observation, timestep, interval, and noisy action are unified as condition-action tokens and processed by a Unified Condition-Action Transformer to predict the interval-aware velocity.
}
\label{fig:overview}
\end{figure*}

\subsection{Overview of UCA-Flow}

As shown in Figure 2, UCA-Flow takes an observation history $o$, a noisy action trajectory $z_t$, a timestep $t$, and an interval length $h$ as inputs, and predicts an interval-aware average velocity $u_\theta(z_t,t,h,o)$ for one-step action generation. It unifies observation, timestep, interval, and action tokens into a single sequence, which is processed by a Unified Condition-Action Transformer to jointly update condition and action representations for dynamic condition reconstruction.

Let $x\in\mathbb{R}^{T\times d_a}$ denote an expert action trajectory and $\epsilon\sim\mathcal{N}(0,I)$ denote Gaussian noise. Following mean-flow, we construct $z_t=(1-t)x+t\epsilon,\ t\in[0,1]$, where larger $t$ indicates a noisier action state, and the instantaneous velocity is $v=\epsilon-x$. UCA-Flow learns $u_\theta(z_t,t,h,o)$ over the interval $h=t-r$. During training, dual-pass supervision optimizes both $u$ and $v$. During inference, starting from $z_1$, it predicts $\hat{x}=z_1-u_\theta(z_1,1,1,o)$ using only the final action-token representations.

\subsection{Unified Condition-Action Tokenization}

As shown on the left of the Figure \ref{fig:overview}, UCA-Flow converts observation tokens, timestep tokens, interval tokens and action tokens into a unified token sequence. The timestep and interval are encoded using sinusoidal embeddings PE followed by independent learnable MLP $\phi$, producing $e_t=\phi_t(\mathrm{PE}(t))$ and $e_h=\phi_h(\mathrm{PE}(h))$. The timestep embedding $e_t$ indicates where the current noisy action lies on the probability path, while the interval embedding $e_h$ specifies the interval over which the average velocity is estimated. To increase representation capacity, we use $N_t$ timestep tokens and $N_h$ interval tokens by adding learnable token-specific biases:
\begin{equation}
\tau_j=e_t+b_j^t,\quad j=1,\dots,N_t,
\end{equation}
\begin{equation}
\eta_j=e_h+b_j^h,\quad j=1,\dots,N_h.
\end{equation}

For observation tokenization, each observation step is represented by a point-cloud feature $p_i\in\mathbb{R}^{d_p}$ and a robot state feature $s_i\in\mathbb{R}^{d_s}$. We concatenate these two features and project them into the Transformer hidden space as $c_i=\phi_o([p_i;s_i])+e_i^o$, with $e_i^o$ serving as a learnable observation-position embedding. With two observation steps, the observation token sequence is $O=[c_1,c_2]$. The condition tokens is then constructed as
\begin{equation}
P_{t,h,o}
=
[c_1,c_2,\tau_1,\dots,\tau_{N_t},\eta_1,\dots,\eta_{N_h}].
\end{equation}

For the noisy action trajectory $z_t$, each horizon step is projected into the same hidden space as $y_k=\phi_a(z_{t,k})+e_k^a,\ k=1,\dots,T$, with $e_k^a$ serving as a learnable action-position embedding. Collecting horizon-wise action tokens gives $Y=[y_1,\dots,y_T]$. Finally, the Transformer input is

\begin{equation}
\begin{aligned}
S^0 &= [P_{t,h,o},Y] \\
    &= [c_1,c_2,\tau_1,\dots,\tau_{N_t},\eta_1,\dots,\eta_{N_h},y_1,\dots,y_T].
\end{aligned}
\end{equation}

\begin{figure}[!h]
  \centering
  \includegraphics[width=\columnwidth]{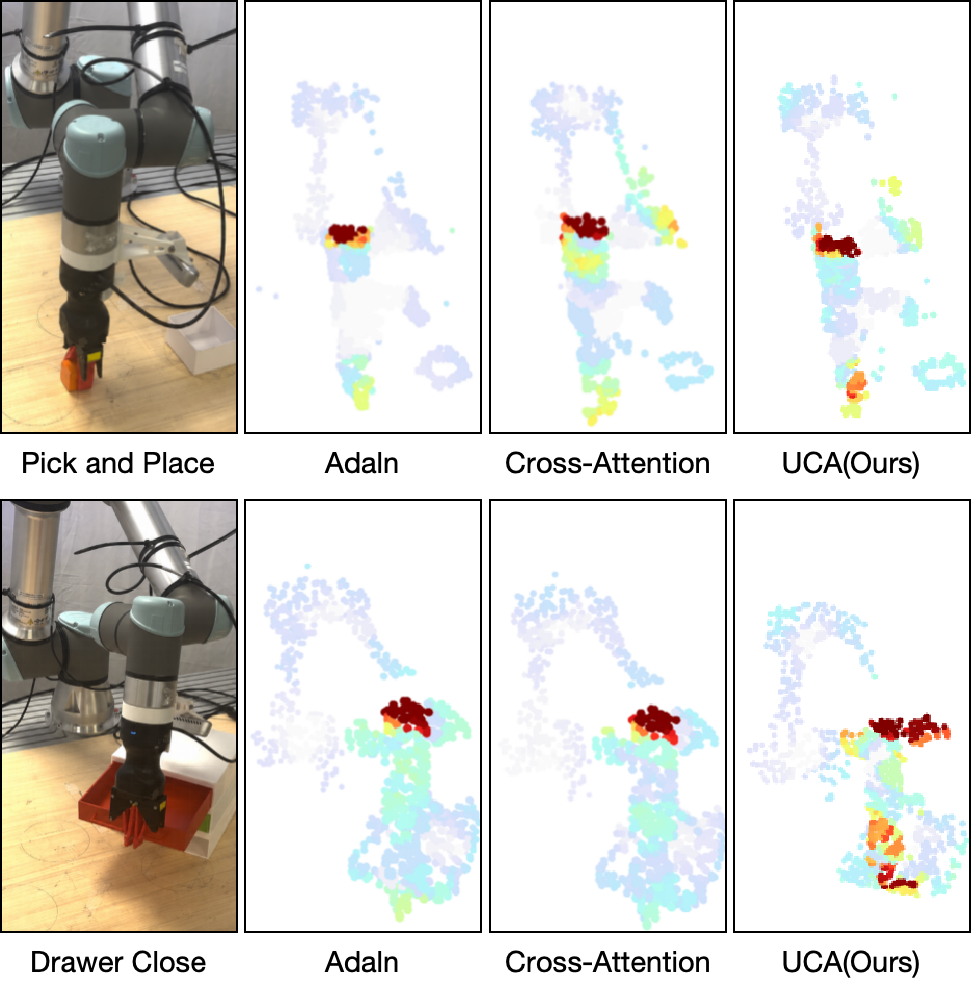}
  \caption{\textbf{Gradient-based saliency visualization on real-world tasks}. UCA-Flow highlights more task-relevant regions, including the end-effector, target object, and upper arm joints, compared with the auxiliary condition-injection AdaLN and Cross-Attention.}
  \label{fig:visual}
\end{figure}

\subsection{Unified Condition-Action Interaction}
As shown on the right of the Figure \ref{fig:overview}, given the unified input sequence $S^0$, UCA-Flow applies a stack of Unified Condition-Action Transformer blocks to jointly update condition tokens and action tokens. For notational clarity, we define the layer-wise condition representations and action representations as
\begin{equation}
P_{t,h,o}^{0}=P_{t,h,o},\qquad Y^0=Y.
\end{equation}
Then each Transformer block jointly updates both condition tokens and action tokens:

\begin{equation}
\begin{aligned}
&[P_{t,h,o}^{\ell+1},Y^{\ell+1}] = \\
&\mathrm{UnifiedConditionActionTransformer}^{\ell}
\left([P_{t,h,o}^{\ell},Y^{\ell}]\right),\\
&\quad \ell=0,\dots,L-1.
\end{aligned}
\end{equation}

Each block consists of multi-head self-attention and a feed-forward network. Unlike designs that update action tokens with externally injected conditions, UCA-Flow processes condition and action tokens within the same Transformer sequence. This allows condition representations to be updated according to the current action state rather than remaining fixed external context, which is also supported by Figure \ref{fig:visual}, where UCA-Flow shows more task-relevant saliency around the end-effector, target object, and upper arm joints.

This creates an action-condition-action information path across layers. 
At layer $\ell$, condition tokens absorb information from the current action tokens $Y^\ell$, forming action-aware condition representations $P_{t,h,o}^{\ell+1}$. 
In subsequent layers, action tokens attend to these updated conditions to refine the trajectory. 
In this way, UCA-Flow turns conditioning from external injection into internal representation evolution, jointly updating condition and action representations during generation.

Finally, we discard the condition outputs and keep only the final action representations:
\begin{equation}
Y^L=[y_1^L,\dots,y_T^L].
\end{equation}
The model prediction is obtained by applying a output head $\psi(\cdot)$ to the action-token sequence:
\begin{equation}
u_\theta(z_t,t,h,o)
=
\psi(Y^L)
\in \mathbb{R}^{T\times d_a}.
\end{equation}

Here $\psi(\cdot)$ denotes a single output head that maps the final action-token sequence to a velocity trajectory with the same horizon length as the action trajectory. Thus, condition tokens are used as dynamic intermediate representations, while the output is produced only from action tokens.

\subsection{Dual-Pass Supervision}

To optimize UCA-Flow for accurate one-step generation, we use an improved dual-pass supervision scheme over the average velocity $u$ and the instantaneous velocity $v$. Given an expert action trajectory $x$ and Gaussian noise $\epsilon$, we sample $t \ge r$ and define $h=t-r$. We then construct $z_t=(1-t)x+t\epsilon$ and $v=\epsilon-x$.
UCA-Flow performs two forward passes during training. The first pass sets $h=0$ and predicts the instantaneous velocity $v_\theta=u_\theta(z_t,t,0,o),$ while the second pass uses the sampled interval length $h$ and predicts the interval-aware average velocity $u_\theta=u_\theta(z_t,t,h,o)$. Following the mean-flow identity, we form a composite velocity by estimating the path-wise derivative of $u_\theta$ with a Jacobian-vector product along $(\mathrm{sg}(v_\theta),\mathbf{1},\mathbf{1})$:
\begin{equation}
D_tu_\theta
=
\nabla_z u_\theta \cdot \mathrm{sg}(v_\theta)
+
\partial_t u_\theta
+
\partial_h u_\theta,
\end{equation}
\begin{equation}
V_\theta
=
u_\theta+h\,\mathrm{sg}(D_tu_\theta).
\end{equation}
Here $\mathrm{sg}(\cdot)$ denotes stop-gradient. We supervise both velocity views with the same target velocity $v$:
\begin{equation}
\mathcal{L}
=
\rho\!\left(V_\theta-\mathrm{sg}(v)\right)
+
\rho\!\left(v_\theta-\mathrm{sg}(v)\right),
\end{equation}
where $\rho(\cdot)$ is the adaptive weighted L2 loss. Given an error tensor $\Delta$, it is defined as
\begin{equation}
\rho(\Delta)
=
\mathrm{sg}\!\left(
\frac{1}{(\|\Delta\|_2^2+c)^{1-\gamma}}
\right)
\|\Delta\|_2^2 .
\end{equation}
Here $\|\cdot\|_2^2$ is computed over the action horizon and action dimensions for each training sample, followed by batch averaging.

The $u$-branch learns interval-aware average velocity for one-step generation, while the $v$-branch provides an auxiliary instantaneous-velocity constraint to stabilize training. During inference, only one forward pass is used: starting from $z_1=\epsilon$, we set $t=1$ and $h=1$, and generate $\hat{x}=z_1-u_\theta(z_1,1,1,o)$.

\begin{table*}[t]
\centering
\caption{\textbf{Ablation study on UCA-Flow components.} 
-DualPass removes dual-pass supervision; 
-Unified/c and Unified/a replace unified condition-action modeling with cross-attention and AdaLN, respectively. 
Results are success rates on representative Adroit and Meta-World tasks.}
\label{tab:core}
\resizebox{\textwidth}{!}{%
\begin{tabular}{c|c|c|c|c|c|c|c|c}
\toprule
\multirow{2}{*}{\textbf{Method}} & \multicolumn{2}{c|}{\textbf{Adroit}} & \multicolumn{6}{c}{\textbf{MetaWorld}} \\ \cmidrule{2-9} 
 & Door & Pen & Dial Turn & Peg Insert Side & Lever Pull & Pick Place Wall & Stick Pull & Shelf Place \\ \midrule
\textbf{UCA-Flow} & $\mathbf{77.3\pm2.1}$ & $\mathbf{63.2\pm3.4}$ & $\mathbf{92.3\pm4.7}$ & $\mathbf{90.7\pm4.0}$ & $\mathbf{89.3\pm3.2}$ & $\mathbf{90.7\pm3.2}$ & $\mathbf{75.3\pm2.3}$ & $\mathbf{74.0\pm3.5}$  \\

-DualPass & $74.3\pm5.0$ & $57.0\pm3.6$ & $91.3\pm7.3$ & $88.0\pm3.6$ & $89.0\pm0.0$ & $89.6\pm5.8 $ & $70.0\pm3.5$ & $64.0\pm7.8$  \\

-Unified/c & $68.2\pm1.8$ & $58.2\pm6.9$ & $87.3\pm10.5$ & $86.3\pm3.2$ & $79.0\pm1.0$ & $84.0\pm3.4 $ & $62.3\pm3.2$ & $56.6\pm3.8$  \\

-Unified/a & $56.5\pm2.1$ & $48.6\pm4.6$ & $78.6\pm9.3$ & $61.0\pm1.7$ & $63.6\pm6.0$ & $60.3\pm11.2$ & $68.0\pm3.5$ & $32.3\pm1.5$  \\ \bottomrule

\end{tabular}%
}
\end{table*}

\begin{figure}[!h]
  \centering
  \includegraphics[width=\columnwidth]{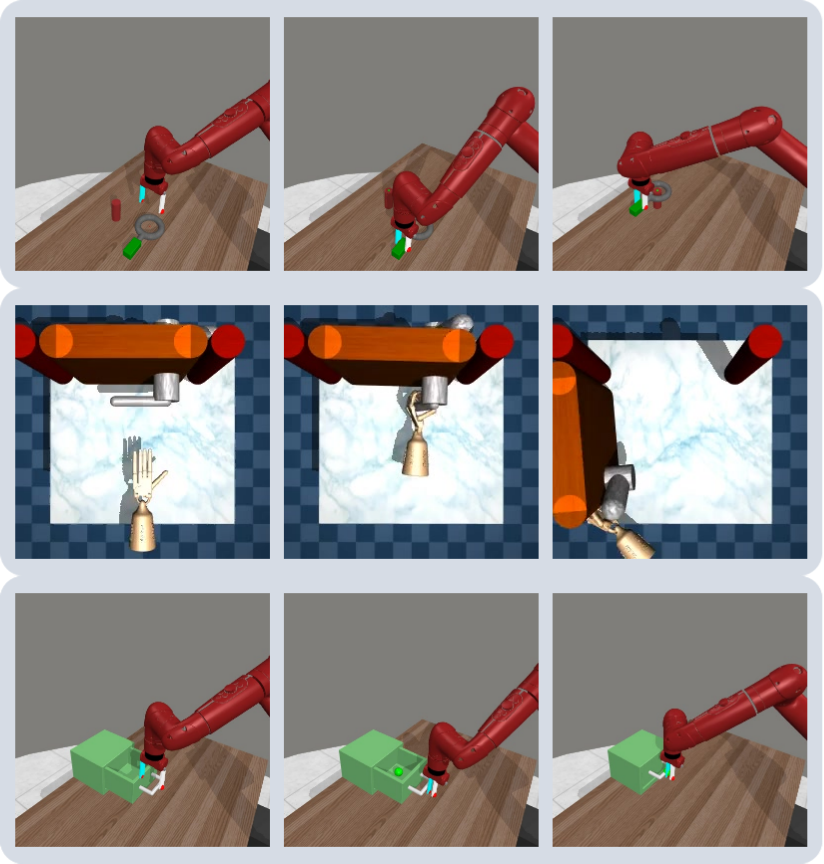}
   \caption{\textbf{Representative simulated rollouts.} Rows correspond to Assembly, Adroit Door, and Drawer Close, and columns show successive stages of each execution. The trajectories require different interaction patterns: aligning and inserting an object, coordinating a dexterous hand with a door handle, and maintaining contact while pushing a drawer closed. UCA-Flow completes all three sequences from approach to task completion using one-step action generation.}
\label{fig:simulated}
\end{figure}

\section{Simulation Experiment}
\subsection{Experiment Setup}

We evaluate UCA-Flow on 37 simulated manipulation tasks from Adroit\cite{rajeswaran2017learning} and Meta-World\cite{yu2020meta}, following prior 3D policy learning works~\cite{ze20243ddiffusionpolicygeneralizable, zhang2025flowpolicy, sheng2025mp1meanflowtamespolicy}.
The benchmark includes 3 Adroit dexterous-hand tasks and 34 Meta-World tabletop tasks across different difficulty levels, 
including 21 Easy, 4 Medium, 4 Hard, and 5 Very Hard tasks. These tasks evaluate action-generation accuracy, task generality, and inference efficiency metrics. We report success rate and inference time as the main performance and efficiency, respectively. All methods use 10 expert demonstrations per task and three seeds (0, 1, 2). 
Point clouds are downsampled by FPS to 1024 points for Adroit and 512 for Meta-World. 
Models are trained for 3,000/1,000 epochs on Adroit/Meta-World, evaluated every 200 epochs, and reported using the top five checkpoints per seed. 
We use AdamW with batch size 128, learning rate $1\times10^{-4}$, observation window 2, and prediction horizon 4 on a single NVIDIA RTX A5000 GPU. 
Inference latency is measured under stable GPU load with three seeds running simultaneously.

\subsection{Ablation Study on the Core Components of UCA-Flow}

\begin{table*}[t]
\centering

\captionof{table}{
\textbf{Success rate comparison on 37 Adroit and Meta-World tasks}.
NFE denotes the number of function evaluations required for action generation.
UCA-Flow group and overall statistics average tasks within each of three seeds before computing the mean and sample standard deviation across seeds; the overall mean weights all 37 tasks equally. Baseline summaries are retained as reported. The best results are highlighted in bold.
}
\label{tab:sota}

\resizebox{\textwidth}{!}{%
\begin{tabular}{l|c|c|ccc|cccc|c}
\toprule
\multirow{2}{*}{Methods} & \multirow{2}{*}{Publication} & \multirow{2}{*}{NFE} & \multicolumn{3}{c|}{Adroit} & \multicolumn{4}{c|}{MetaWorld} & \multirow{2}{*}{\textbf{Average}} \\ 
 & & & Hammer & Door & Pen & Easy (21) & Medium (4) & Hard (4) & Very Hard (5) & \\ \midrule

DP3 & RSS'24 & 10 & $100\pm0$ & $56\pm5$ & $46\pm10$ & $87.3\pm2.2$ & $44.5\pm8.7$ & $32.7\pm7.7$ & $39.4\pm9.0$ & $68.7\pm4.7$ \\
Simple DP3 & RSS'24 & 10 & $98\pm2$ & $40\pm17$ & $36\pm4$ & $86.8\pm2.3$ & $42.0\pm6.5$ & $38.7\pm7.5$ & $35.0\pm11.6$ & $67.4\pm5.0$ \\
FlowPolicy & AAAI'25 & 1 & $98\pm1$ & $61\pm2$ & $54\pm4$ & $84.8\pm2.2$ & $58.2\pm7.9$ & $40.2\pm4.5$ & $52.2\pm5.0$ & $71.6\pm3.5$ \\ 
MP1 & AAAI'26 & 1 & $100\pm0$ & $69\pm2$ & $58\pm5$ & $88.2\pm1.1$ & $68.0\pm3.1$ & $58.1\pm5.0$ & $67.2\pm2.7$ & $78.9\pm2.1$ \\\midrule
\textbf{UCA-Flow} & \textbf{Ours} & 1 & $\bm{100\pm0}$ & $\bm{77.3\pm2.1}$ & $\bm{63.2\pm3.4}$ & $\bm{92.4\pm0.7}$ & $\bm{79.3\pm1.8}$ & $\bm{68.0\pm1.9}$ & $\bm{85.3\pm0.7}$ & $\bm{86.4\pm0.3}$ \\ \bottomrule
\end{tabular}%
}

\captionof{table}{
\textbf{Inference time comparison on Adroit and MetaWorld}.
We report inference latency in milliseconds with the reported standard deviation. Overall mean latency weights task groups by their task counts; speedup ratios are computed before rounding the overall means.
NFE denotes the number of function evaluations required.
All methods are evaluated with three random seeds, and the best results are highlighted in bold.
}
\label{tab:speed}

\resizebox{\textwidth}{!}{%
\begin{tabular}{l|c|c|ccc|cccc|c}
\toprule
\multirow{2}{*}{Methods} & \multirow{2}{*}{Publication} & \multirow{2}{*}{NFE} & \multicolumn{3}{c|}{Adroit /ms} & \multicolumn{4}{c|}{MetaWorld /ms} & \multirow{2}{*}{\textbf{Average /ms}} \\ 
 & & & Hammer & Door & Pen & Easy (21) & Medium (4) & Hard (4) & Very Hard (5) & \\ \midrule
DP3 & RSS'24 & 10 & $129.5\pm13.9$ & $141.3\pm14.8$ & $145.1\pm12.3$ & $129.3\pm10.7$ & $134.7\pm11.5$ & $131.9\pm12.4$ & $138.4\pm10.8$ & $132.2\pm11.2$ \\
Simple DP3 & RSS'24 & 10 & $103.1\pm11.4$ & $111.3\pm10.2$ & $128.2\pm13.1$ & $91.9\pm8.6$ & $98.3\pm9.1$ & $101.3\pm9.7$ & $103.8\pm10.2$ & $97.0\pm9.2$ \\
FlowPolicy & AAAI'25 & 1 & $15.3\pm1.1$ & $13.2\pm4.0$ & $12.0\pm2.8$ & $12.0\pm1.4$ & $12.2\pm1.5$ & $13.5\pm1.4$ & $14.5\pm1.6$ & $12.6\pm1.5$ \\ 
MP1 & AAAI'26 & 1 & $7.1\pm0.2$ & $7.2\pm0.1$ & $7.4\pm0.3$ & $6.7\pm0.0$ & $6.7\pm0.1$ & $6.7\pm0.1$ & $6.8\pm0.1$ & $6.8\pm0.1$ \\ \midrule
\textbf{UCA-Flow} & \textbf{Ours} & 1 & $\bm{2.9\pm0.2}$ & $\bm{2.9\pm0.1}$ & $\bm{3.0\pm0.3}$ & $\bm{3.1\pm0.1}$ & $\bm{2.9\pm0.1}$ & $\bm{3.0\pm0.1}$ & $\bm{2.8\pm0.1}$ & $\bm{3.0\pm0.1}$ \\ \bottomrule
\end{tabular}%
}

\end{table*}

Table~\ref{tab:core} shows that the full UCA-Flow performs best across most representative tasks. Removing dual-pass supervision causes moderate drops, e.g., Pen decreases from 63.2\% to 57.0\% and Shelf Place from 74.0\% to 64.0\%, confirming the benefit of complementary $u$ and $v$ supervision. Replacing unified condition-action modeling with cross-attention injection leads to larger drops, such as Shelf Place from 74.0\% to 56.6\%. AdaLN-based injection performs worst, with clear failures on Door (77.3\% → 56.5\%), Peg Insert Side (90.7\% → 61.0\%), and Shelf Place (74.0\% → 32.3\%). These results support our motivation that conditions should be jointly modeled with action tokens rather than externally injected as auxiliary signals.

\subsection{Comparison with State-of-art Methods}
\paragraph{Performance and speed comparison.}
Table~\ref{tab:sota} and Table~\ref{tab:speed} compare UCA-Flow with representative diffusion-based and flow-based policies.
UCA-Flow achieves the best average success rate of $86.4\pm0.3\%$, improving over the reported MP1 mean of $78.9\%$ by $7.5$ percentage points and outperforming the one-step FlowPolicy by $14.8$ percentage points.
The gains are especially clear on challenging Meta-World tasks, where UCA-Flow reaches $79.3\%$, $68.0\%$, and $85.3\%$ on Medium, Hard, and Very Hard tasks, respectively. UCA-Flow also achieves the lowest task-averaged inference latency of $3.0$ ms with only one function evaluation.
Using task-weighted latencies before rounding, it achieves a $43.9\times$ speedup over DP3, $32.2\times$ over Simple DP3, $4.2\times$ over FlowPolicy, and $2.2\times$ over MP1.
These results demonstrate that unified condition-action modeling enables accurate and real-time action generation with a compact model.

\begin{figure}[!h]
  \centering
  \includegraphics[width=\columnwidth]{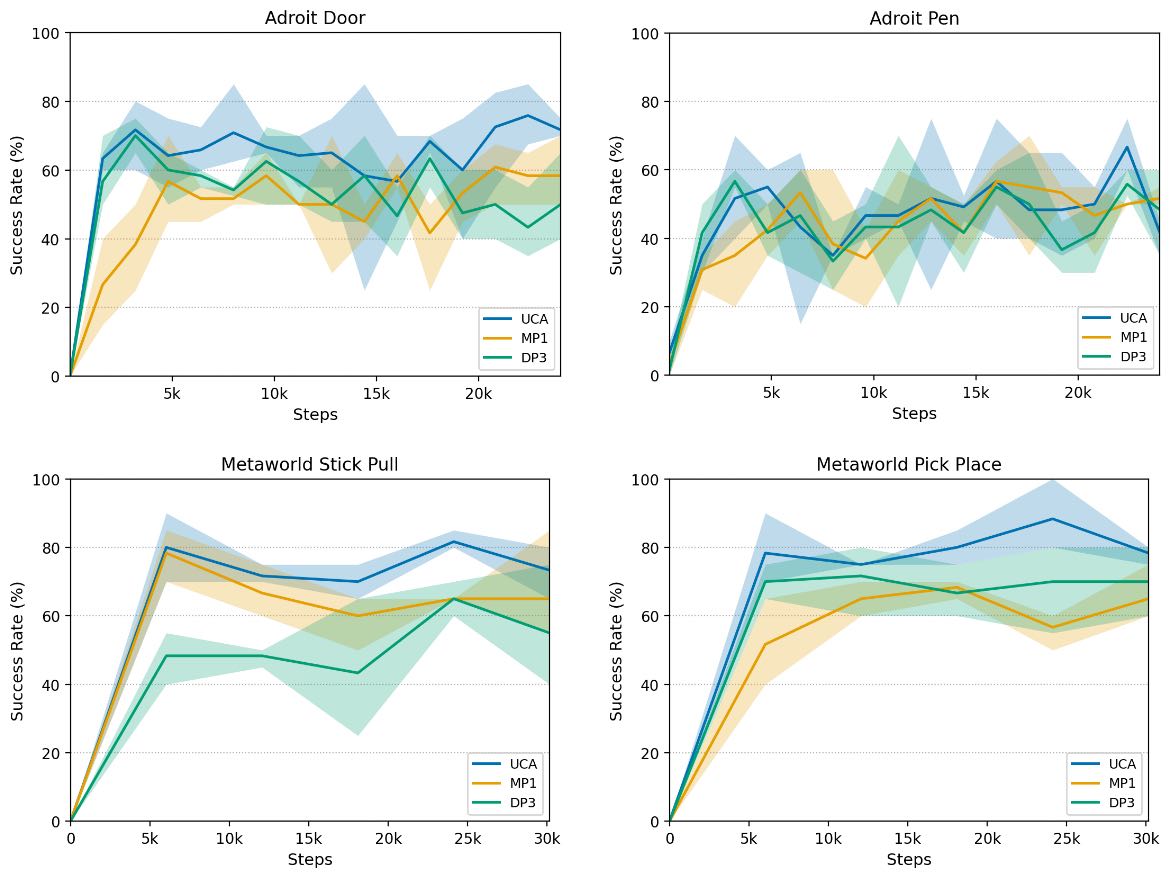}
  \caption{\textbf{Training success curves on Adroit and Meta-World.} Solid lines show success rates and shaded regions indicate variation across seeds. Using the top-five-checkpoint mean per seed, UCA-Flow outperforms the baselines on all four tasks, including an approximately 5\% gain on Adroit Pen.}
\label{fig:simulated_curves}
\end{figure}

\begin{figure*}
  \centering
  \includegraphics[width=\textwidth]{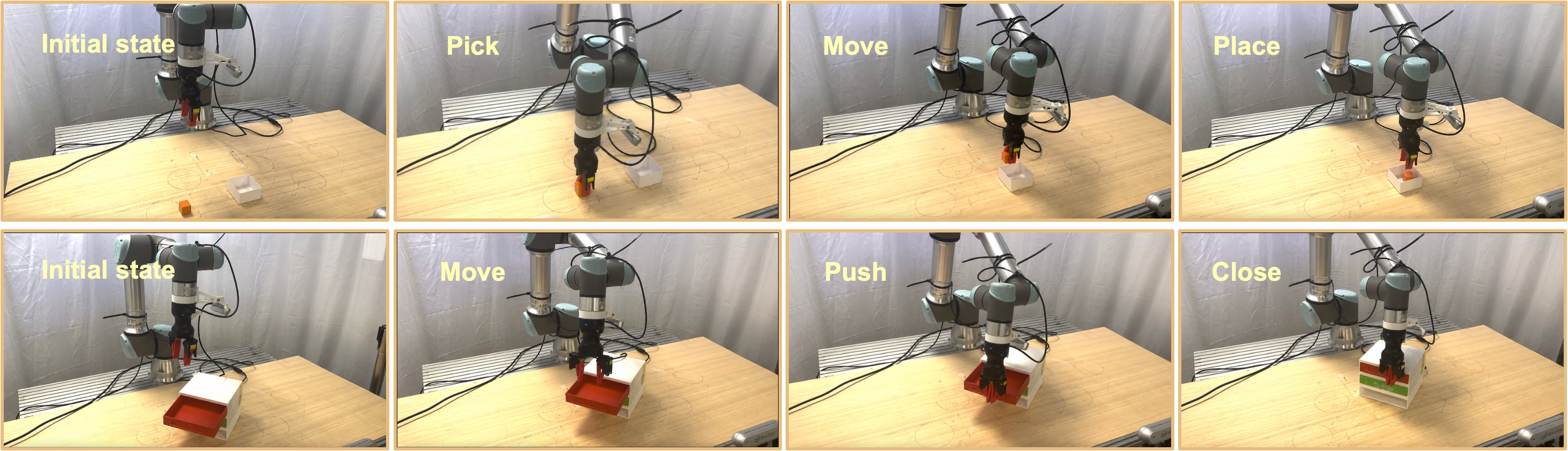}
\caption{
\textbf{Real-world execution sequences.} The top row shows Pick and Place progressing from the initial scene through grasping and transport to placement in the target box. The bottom row shows Drawer Close progressing from approach to contact, sustained pushing, and full closure. These sequences illustrate that UCA-Flow maintains task-relevant control through distinct free-space and contact-rich phases despite sensing and execution uncertainty.
}
\label{fig:realworld}
\end{figure*}
\section{Realworld Experiments}

\subsection{Experiment Setup}

We conduct real-world experiments on a UR5e robotic arm with a RealSense L515 RGB-D camera mounted for a top-down workspace view, as shown in Figure \ref{fig:realworld}. We evaluate two tasks: Pick and Place and Drawer Close, involving object placement, drawer closing under different opening degrees. For each task, we collect 40 human demonstrations and use a unified setting with prediction horizon 16 and observation stride 2. Models are trained for 300 epochs per task . We use the AdamW optimizer with a batch size of 128 and learning rate $  1 \times 10^{-4}  $. Each task is evaluated over 20 independent trials.

\subsection{Realworld Experimental Results}
Figure \ref{fig:realworld} visualizes the real-world robot tasks, including pick-and-place and drawer-closing. The sequential snapshots show that UCA-Flow can generate executable actions across different manipulation stages, such as approaching the object, completing the grasp, moving to the target region, and closing the drawer. Table \ref{tab:realworld_main} compares UCA-Flow with representative baselines on these two real-world tasks, where UCA-Flow improves Pick and Place from 55.0\% to 65.0\% over MP1 and reaches 100.0\% success on Drawer Close. Table \ref{tab:realabl}  further evaluates key design choices in real-world settings. Compared with AdaLN and Cross Attention, UCA-Flow achieves higher success rates on both tasks, indicating that unified condition-action modeling is more effective than auxiliary condition injection. These results suggest that UCA-Flow remains effective under real-world sensing noise and execution uncertainty, providing evidence for its real-world manipulation capability.

\begin{table}[htbp]
  \centering
  \small
  \caption{\textbf{Real-world comparison with policy baselines.} Success rates (\%) are measured over 20 independent trials per task. UCA-Flow improves Pick and Place by 10 percentage points over MP1 and achieves 100\% success on Drawer Close.}
  \label{tab:realworld_main}
  \begin{tabular}{l|cc}
    \toprule
    \textbf{Method} & Pick \& Place & Drawer Close \\
    \midrule
    DP3                & 50.0  & 95.0  \\
    MP1                & 55.0  & 95.0  \\
    \textbf{UCA-Flow}  & \textbf{65.0} & \textbf{100.0} \\
    \bottomrule
  \end{tabular}
\end{table}

\begin{table}[htbp]
  \centering
  \small
  \caption{\textbf{Real-world comparison of condition-integration designs.} Success rates (\%) are measured over 20 independent trials per task. Unified condition-action modeling performs best on both tasks, exceeding AdaLN by 5 and 10 percentage points on Pick and Place and Drawer Close, respectively.}
  \label{tab:realabl}
  \begin{tabular}{l|cc}
    \toprule
    \textbf{Method} & Pick \& Place & Drawer Close \\
    \midrule
    AdaLN            & 60.0 & 90.0 \\
    Cross Attention  & 50.0 & 95.0 \\
    \textbf{UCA-Flow} & \textbf{65.0} & \textbf{100.0} \\
    \bottomrule
  \end{tabular}
\end{table}

\section{Conclusion}
In this paper, we proposed UCA-Flow, a unified condition-action modeling framework for accurate and efficient one-step action generation. UCA-Flow represents observation, timestep, interval, and action tokens in a unified sequence and updates them with a shared Transformer, enabling condition and action representations to be jointly refined during generation. We further introduced dual-pass supervision to improve one-step velocity prediction. Experiments on Adroit and Meta-World show that UCA-Flow outperforms representative diffusion- and flow-based policies while maintaining low inference latency. Real-world experiments further validate its effectiveness on physical robot tasks, demonstrating that unified condition-action modeling is a simple and effective design for fast generative robot policies.
\section{Limitation}
UCA-Flow mainly focuses on accurate and efficient one-step action generation, and our current real-world evaluation is conducted on representative tabletop manipulation tasks. Although these experiments validate the effectiveness of unified condition-action modeling in physical robot settings, future work can further extend UCA-Flow to broader real-world scenarios, such as longer-horizon manipulation and more contact-rich tasks.

\clearpage
\newpage
\bibliographystyle{assets/plainnat}
\bibliography{paper}

\end{document}